\documentclass{article}
\usepackage{main}

\usepackage{times}
\usepackage{soul}
\usepackage{url}
\usepackage[hidelinks]{hyperref}
\usepackage[utf8]{inputenc}
\usepackage[small]{caption}
\usepackage{graphicx}
\usepackage{amsmath}
\usepackage{amsthm}
\usepackage{booktabs}
\usepackage{algorithm}
\usepackage{algpseudocode}
\usepackage[switch]{lineno}

\usepackage{tablefootnote}
\usepackage{enumitem}
\usepackage{amsfonts}

\usepackage{mathtools}
\DeclarePairedDelimiter\ceil{\lceil}{\rceil}

\title{Synthesizing Composite Hierarchical Structure from Symbolic Music Corpora}

\author{
Ilana Shapiro$^1$
\and
Ruanqianqian (Lisa) Huang$^1$\and
Zachary Novack$^1$\and
Cheng-i Wang$^2$\and
Hao-Wen Dong$^1$\and
Taylor Berg-Kirkpatrick$^1$\and
Shlomo Dubnov$^1$\And
Sorin Lerner$^1$
\\
\affiliations
$^1$University of California, San Diego, CA, USA\\
$^2$AudioShake, Oakland, CA, USA
\emails
\{ilshapiro, r6huang, znovack, h3dong, sdubnov, tberg, lerner\}@ucsd.edu,
\{chw160\}@audioshake.ai
}

\begin{document}

\maketitle

\begin{abstract}
Western music is an innately hierarchical system of interacting levels of structure, from fine-grained melody to high-level form. In order to analyze music compositions holistically and at multiple granularities, we propose a unified, hierarchical meta-representation of musical structure called the \textit{structural temporal graph} (STG). For a single piece, the STG is a data structure that defines a hierarchy of progressively finer structural musical features and the temporal relationships between them. We use the STG to enable a novel approach for deriving a representative structural summary of a music corpus, which we formalize as a nested NP-hard combinatorial optimization problem extending the Generalized Median Graph problem. Our approach first applies simulated annealing to develop a measure of \emph{structural distance} between two music pieces rooted in graph isomorphism. Our approach then combines the formal guarantees of SMT solvers with nested simulated annealing over structural distances to produce a structurally sound, representative \emph{centroid} STG for an entire corpus of STGs from individual pieces. To evaluate our approach, we conduct experiments verifying that structural distance accurately differentiates between music pieces, and that derived centroids accurately structurally characterize their corpora.
\end{abstract}

\section{Introduction}\label{sec:intro}
A prevailing theory among Western music theorists and musicologists states that Western classical music exhibits an implicitly hierarchical structure~\cite{simonetta_article}. While several different theoretical systems have been proposed to formalize this structural hierarchy~\cite{marsden2013}, a widely accepted modern interpretation of the hierarchy states that melodies form the bottom, followed by harmonic contour, rhythmic patterns, disjoint and possibly overlapping motifs, and finally contiguous sections~\cite{msaf,milne,dai2024}. Together, this composite hierarchy encapsulates the overall structure of a piece. 

To analyze musical structure computationally, many automatic approaches have been developed for extracting structure at \emph{single} levels of the structural hierarchy \cite{Hsiao_2023_motifs,rhythm_extraction,harmonytransformerV2,salamon_2013_melody}, including methods to analyze sub-hierarchies within a single level \cite{eval_hier}. However, music perception researchers have shown that the levels are not perceptually independent: they relate to one another in both ``vertical" (structural) and ``horizontal" (temporal) directions~\cite{Narmour_1983}, interactions that have more recently been proven computationally ~\cite{dai2024}. A comprehensive analysis of a piece thus must integrate these inter- and intra-level interactions into a unified model. Furthermore, while attempts have been made to \emph{generate} structured music \cite{young_2022,hyperbolic_hierarchy,wang2024wholesong}, to our knowledge, no existing research has examined generating music adhering to a complete, musically representative structural hierarchy, since no mechanism to computationally derive a complete structural specification from a desired music corpus currently exists. Such an encapsulation of music structure could also play a critical role in the generation of well-formed music by serving as a system of constraints on generative models. 

Despite prior attempts at such an integrated, computational model of musical form, two challenges remain. First, prior approaches do not completely encapsulate the vertical and horizontal relationships of the structural hierarchy, cannot handle polyphonic music, or are not fully automatic~\cite{gttm_implementation,simonetta_article,mokbel2009_graph,music_hier_review}. Second, to our knowledge, existing methodologies focus only on \textit{individual} pieces with no attempt to \textit{summarize} the hierarchy over a music corpus to synthesize its overall structure and obtain a holistic representation of the entirety of the corpus. 

To address these challenges, we introduce the \textit{structural temporal graph} (STG) as a unified model of complete musical structure. The STG is a $k$-partite directed acyclic graph whose levels form the structural hierarchy, and edges encode temporal relationships between adjacent levels. We use simulated annealing to develop a measure of \emph{structural distance} between two STGs based on graph isomorphism, and to obtain the overarching structure of a corpus of pieces, we develop an approach to derive a representative \emph{centroid} graph from a corpus of STGs. We formalize centroid derivation as a nested NP-hard combinatorial optimization problem extending the Generalized Median Graph problem \cite{Jiang2001}, and propose a solution combining nested simulated annealing with the formal guarantees of SMT solvers to produce a structurally sound result. Our experiments show that structural distance accurately differentiates pieces, with its performance reliant on the complete hierarchy, and that derived centroids accurately structurally characterize their corpora.\footnote{Paper code: \url{https://github.com/ilanashapiro/stg_optimization}}

In summary, the contributions of this paper are as follows:
\begin{enumerate}
\item We propose the \textit{structural temporal graph}, a meta-representation of musical form unifying the entire structural hierarchy, and develop a \textit{structural distance} measure between two STGs rooted in graph isomorphism. 

\item We formalize the music summarization problem as a nested NP-hard combinatorial optimization problem, and contribute a novel solution using both stochastic and SMT-based techniques.

\item We conduct experiments verifying structural distance accurately differentiate pieces, and music corpora are accurately characterized by their derived centroids.

\end{enumerate}
\section{Related Work}\label{sec:related_work}

\noindent\textbf{Single-Level Analyses.}
Many existing algorithms extract structure at \textit{single} levels of the music structural hierarchy. To extract segmentation, The Music Structure Analysis Framework (MSAF) toolkit~\cite{msaf} features factorization-based techniques, including ordinal linear discriminant analysis~\cite{olda}, convex nonnegative matrix factorization~\cite{cnmf}, checkerboard~\cite{checkerboard}, spectral clustering~\cite{sCluster}, the Structural Features algorithm~\cite{sf}, 2D-Fourier Magnitude Coefficients~\cite{2dfmc}, and the Variable Markov Oracle~\cite{vmo_segmentation}. Motif discovery algorithms search for disjoint, repeating, and possibly overlapping patterns in a piece. String-based approaches~\cite{vmo_motifs} represent music as a chromagram and detect patterns with sub-string matching, and geometry-based approaches~\cite{Hsiao_2023_motifs} represent music as multidimensional point sets, and translatable subsets identify patterns. Recent approaches in harmony identification are centered around neural networks, such as using transformers to incorporate chord segmentation into the recognition process~\cite{chen_2019_harmony,harmonytransformerV2}. Until very recently, the Melodia algorithm was the state of the art in melody extraction, but recent approaches have shifted to neural networks~\cite{kosta_22_melody,midibert}. 

\smallskip\noindent\textbf{Integrated Models of Structure.}
Music theorists have attempted to unify the structural hierarchy with frameworks such as Schenkerian theory~\cite{schenkerian} and the Generative Theory of Tonal Music (GTTM)~\cite{gttm}. Schenkerian analysis applies a series of reductions that progressively simplify a musical piece by removing layers of structure. Attempts to automatically derive Scherkerian analyses are intractable for all but very short pieces, and have low accuracy~\cite{schenkerian}. GTTM generates four different structural hierarchies (grouping structure, metrical structure, time-span tree, and prolongational tree) for a piece of music, to model human cognition~\cite{gttm_implementation}. Computational implementations GTTM (e.g. the Automatic Timespan Tree Analyser~\cite{gttm_implementation}) cannot handle polyphonic music, and are not fully automatic. Improved results with these theories are unlikely, as neither gives the precision required for complete computational implementation~\cite{marsden2013}.

Such theoretical limitations led to a modern interpretation of the structural hierarchy: segmentation, motifs (disjoint/repeating patterns), rhythm, harmony, and melody~\cite{msaf,milne,dai2024}. Many approaches partially encode the hierarchy in graphs: topographic mappings for melodic progressions ~\cite{mokbel2009_graph}, graphs for interactions between sections, melody, harmony and rhythm ~\cite{cmu_dannenberg_2020}, multi-edge graphs for bar-level relations~\cite{music_structure_lit_review}, and undirected graphs for melodies and their reductions ~\cite{orio2009}. The prototype graph~\cite{young_2022} is a bipartite network relating prototype elements to the music they represent. Attempts to model the structural hierarchy with formal grammars~\cite{smallest_grammar,repetition_grammars_ismir2023} are limited to segmentation and motifs.

None of these approaches encapsulate the entire hierarchy, and to our knowledge, there have also been no attempts to synthesize representative structure from a music corpus. 

\section{Structural Temporal Graph}\label{sec:stg}
To address the lack of a fully automatic complete encapsulation of polyphonic musical structure, we introduce the \textit{structural temporal graph} (STG), a unified meta-representation of musical structure that captures the levels of the music structural hierarchy and the temporal relationships between them. The STG is a $k$-partite directed acyclic graph (DAG), where each of the $k$ layers encodes a level in the music structural hierarchy.\footnote{Individual levels themselves can form sub-hierarchies of increasing granularity, which the STG supports} Following the modern music theoretic interpretation of the hierarchy~\cite{msaf,milne,dai2024}, from top to bottom we denote the levels to be contiguous segmentation, motifs (both disjoint and overlapping), rhythmic contour, relative keys, functional harmonic chords, and melodic contour. The STGs we build include every level in the hierarchy except rhythmic contour, for which we were unable to access an analysis algorithm. We run individual analysis algorithms to generate each level of the hierarchy, which is elucidated in Section \ref{sec:eval}. Before formally defining the STG, we build intuition by walking through the derivation of an STG from an annotated piece. 

\begin{figure}[t]
  \centering
  \includegraphics[width=0.9\linewidth]{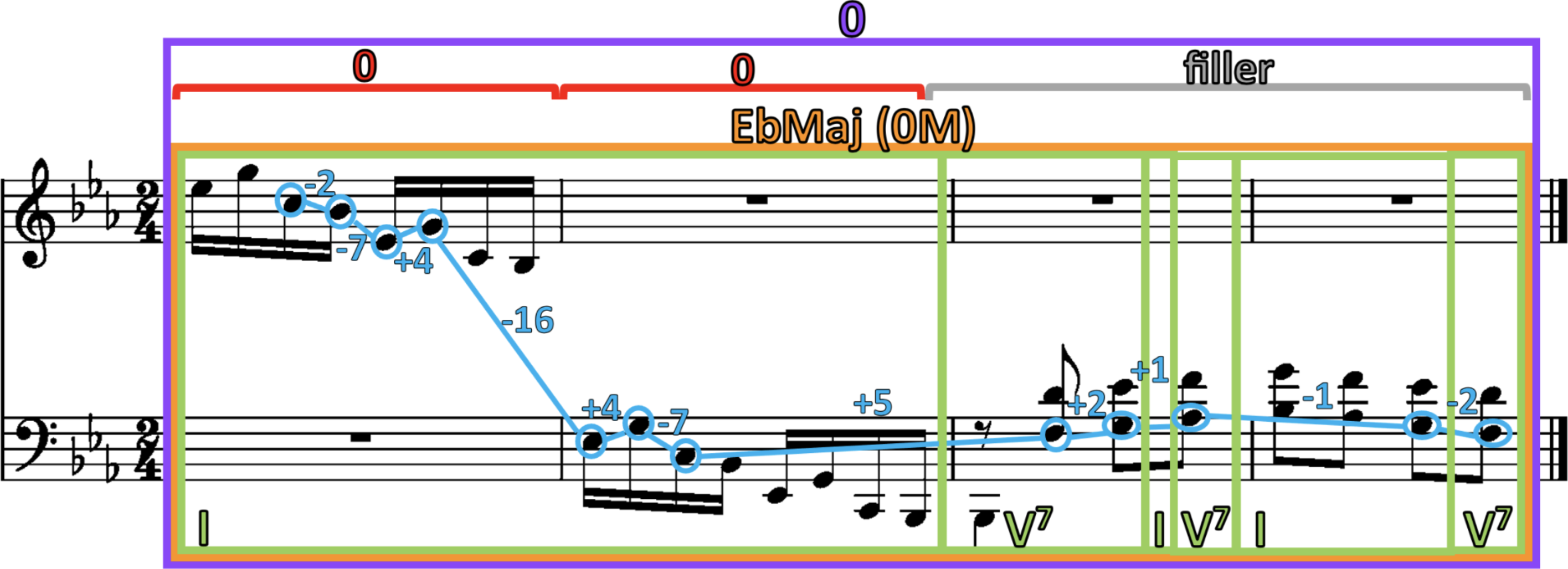}
  \caption{Computer generated analysis of Beethoven's Biamonti Sketch No. 461.}
  \label{fig:beethoven_score_markedup}
\end{figure}

\smallskip\noindent\textbf{Building the Graph.}
We walk through the derivation of an STG from Beethoven's Biamonti Sketch No. 461 that unifies contiguous segmentation; disjoint, overlapping motifs; relative keys; functional harmonic chords; and melodic contour. First, we manually analyze the piece by annotating its score with computer-generated hierarchical structure analyses in Figure~\ref{fig:beethoven_score_markedup}.
Each colored annotation corresponds to one level of the structural hierarchy. 
In purple, we see that this piece has one large contiguous segment, labeled 0. Next, disjoint motifs are in red. Motif 0 appears twice, at the beginning of bars 1 and 2. The gray filler bar indicates no more motifs appear in the latter half. In orange, we see this piece is in a single key, Eb Major. We label this key symbolically as 0M to indicate this is relative key number 0 (i.e. the first key) in M for major. Subsequent keys would be numbered by their positive interval difference from the previous key within the 12-tone scale. We next see functional harmonic chords in green annotated with Roman numeral chord symbols, and finally melodic contour intervals in blue. Notably, such generated analyses may be slightly inaccurate (e.g. the third green I chord should correspond to the previous beat). Since the STG is a fully automatic \textit{meta}-representation of musical structure, it is only as accurate as the analysis algorithms it uses.

\begin{figure}[t]
  \centering
  \includegraphics[width=0.9\linewidth]{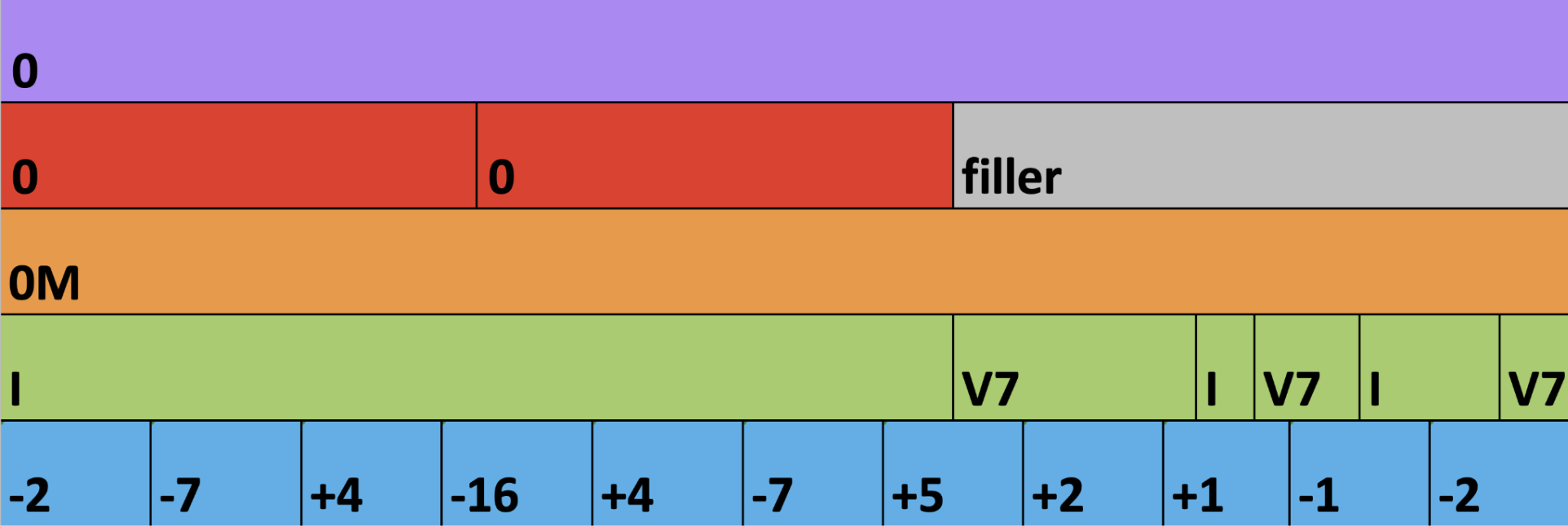}
  \caption{Spatial Visualization of Figure \ref{fig:beethoven_score_markedup}}
  \label{fig:beethoven_generated_rectangles}
\end{figure}

\begin{figure}[t]
  \centering
  \includegraphics[width=0.9\linewidth]{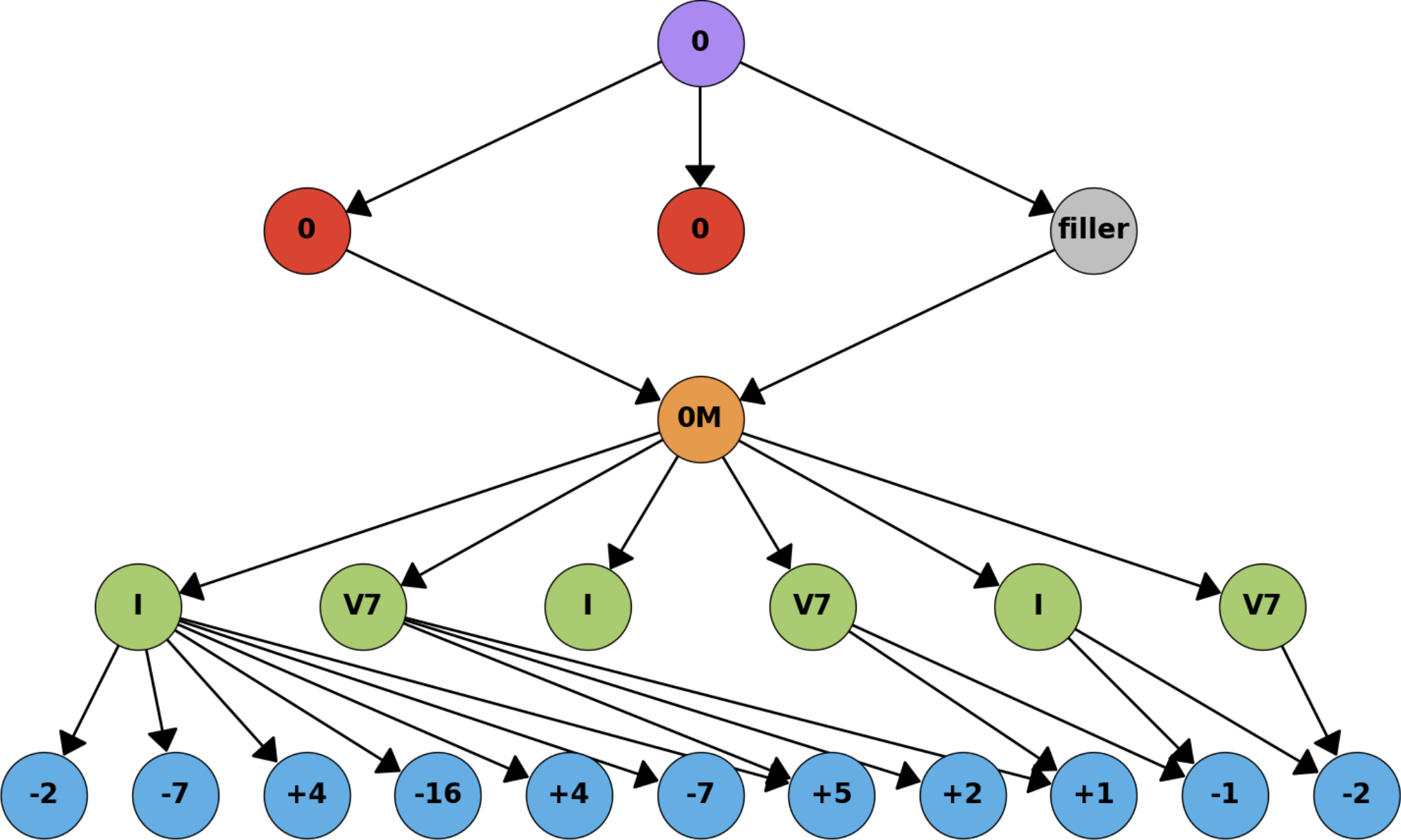}
  \caption{STG for Beethoven's Biamonti Sketch No. 461}
  \label{fig:beethoven_stg}
\end{figure}

We then equivalently represent the ground-truth annotations in Figure \ref{fig:beethoven_score_markedup} as the stacked rectangles in Figure \ref{fig:beethoven_generated_rectangles} to elucidate how each level of the structural hierarchy relates to the next. Finally, we transition to the STG in Figure \ref{fig:beethoven_stg}. There is a surjection between Figures \ref{fig:beethoven_score_markedup}, \ref{fig:beethoven_generated_rectangles}, and \ref{fig:beethoven_stg}. The edges and nodes of the STG, respectively, correspond to the vertical and horizontal alignments of the rectangles in Figure \ref{fig:beethoven_generated_rectangles}. All motif nodes, including the gray filler node indicating no motifs for that interval, fall into the time interval of purple segmentation node 0. The orange key node 0M \emph{starts} in the first red motif node 0, and \emph{ends} in the last gray motif filler node (i.e. the key spans the entire piece). All the green chord nodes fall in the orange key node's interval. Finally, we see how blue melodic contour nodes relate to green chord nodes. For instance, the first melody interval -2 begins and ends in the first chord node I, and the penultimate melody interval -1 begins in the fourth V7 chord node and ends in the fifth I chord node.

Formally, the nodes of an STG encode labeled musical sections along with their associated time intervals generated by the relevant analysis algorithm. Nodes are sorted within each level based on start time, and edges encode temporal relationships between nodes of adjacent levels. Specifically, for node $n$ at level $i$, $n$ must have either one or two parents in level $i-1$ directly above it: one if its associated time interval is a total subset of its parent's, and two if its time interval begins in one parent's, and ends in the other's. 

\section{Structural Distance}\label{sec:struct_dist}
At a high level, the distance between two STGs is the minimum number of edit operations (deletion, insertion, and substitution of nodes/edges) required to transform one graph to the other, also known as graph edit distance (GED)~\cite{ged}. 
However, GED measures \textit{isomorphic} similarity between two graphs, i.e. it evaluates how closely graph structures match independent of labeling. We cannot currently leverage STG isomorphism because STGs are ``compressed," with structure encoded in node ids. Specifically, structure is encoded in the defining features of each node id, and in the intra-level linear temporal orderings for each analysis (i.e. the horizontal order of each level in the graph, currently determined by node index). Thus, in order to reason about STGs isomorphically, we must augment them to encode all structural attributes directly within the graph's topology.

\begin{figure}[t]
  \centering
  \includegraphics[width=0.8\linewidth]{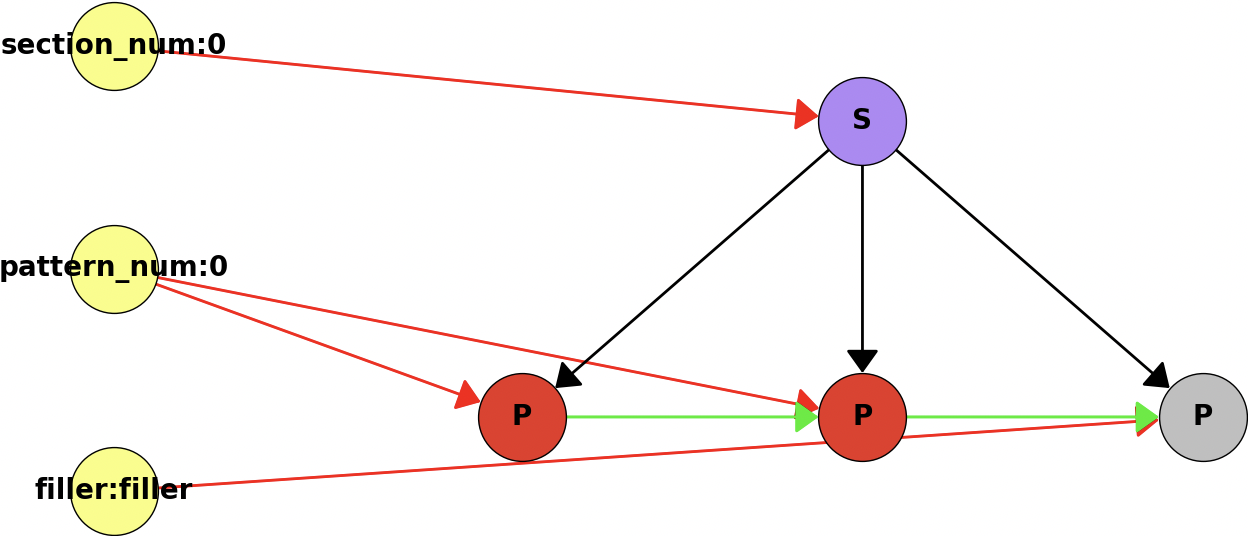}
  \caption{Augmenting the first two levels of the Beethoven STG. Yellow prototype nodes have the format feature\_name:feature\_value}
  \label{fig:beethoven_stg_aug_zoom}
\end{figure}

To encode element labels, recall that each node id encodes a defining feature set. All nodes can thus be alternatively encoded as \emph{instances} of their feature \emph{prototypes}. We create a \emph{prototype node} for each feature and assign it as a parent of the corresponding instance node(s) with that feature. For instance, segmentation nodes encode a single feature: the section number they correspond to. Finally, to encode intra-level linear temporal relationships, we form a linear chain with edges between pairs of horizontally adjacent nodes. This results in a structurally complete STG we can reason about isomorphically. Figure \ref{fig:beethoven_stg_aug_zoom} shows the first two levels of the STG from Figure \ref{fig:beethoven_stg}, with yellow prototype nodes on the left for each instance feature (section number for segmentation nodes \textbf{S}, and pattern number and filler for motif/pattern nodes \textbf{P}), red edges connecting prototype features to instance nodes, and green edges for the pattern layer intra-level linear chain.

\subsection{Graph Alignment Annealing}\label{subsec:SA_align}
GED is a NP-hard combinatorial optimization problem~\cite{ged}, intractable for most STGs. Most GED approximation algorithms are slow and of dubious accuracy, and more generalized than we require \cite{ged_acc}. We thus introduce a new measure of \emph{structural distance} computed with simulated annealing (SA), a stochastic optimization technique that estimates the global optimum of a discrete cost function. It comprises an objective ``energy" function to minimize and a ``move" function for generating a new solution from the current state. An annealer begins at a high \emph{temperature} indicating the likelihood of accepting worse solutions to explore the solution space, and ends at a low temperature to refine near-optimal solutions \cite{simanneal}.

To use SA, we convert the augmented STGs to adjacency matrices, and pad each matrix with zero-arity ``dummy nodes" so they have identical dimensions. Given such matrices $A_{G_1}$ and $A_{G_2}$, their distance is given in Equation \ref{eq:dist}, where \( \left\| . \right\|_F \) denotes the Frobenius norm. When $A_{G_1}$ and $A_{G_2}$ are optimally aligned, $\textsc{dist}(A_{G_1}, A_{G_2})$ is simply $\sqrt{\text{GED}}$.
\begin{equation}
\label{eq:dist}
\textsc{dist}(A_{G_1}, A_{G_2}) = \left\| A_{G_1} - A_{G_2} \right\|_F
\end{equation} 
Finding permutation matrix $P$ optimally aligning $A_{G_2}$ to $A_{G_1}$ to minimize Equation \ref{eq:dist} is NP-hard, so we use SA. The alignment annealer's energy function is given in Equation \ref{eq:alignment_annealer_energy}. Given optimal $P$, Equation \ref{eq:alignment_annealer_energy} computes the structural distance between $A_{G_1}$ and $A_{G_2}$.
\begin{equation}
\label{eq:alignment_annealer_energy}
\textsc{energy}(A_{G_1}, A_{G_2}, P) = \textsc{dist}(A_{G_1}, P^T A_{G_2} P)
\end{equation}

The move function for modifying $P$ at each step of SA is given in Algorithm \ref{alg:alignment_annealer_move}.
\begin{algorithm}[t]
\caption{Alignment Annealer Move Function}
\label{alg:alignment_annealer_move}
\begin{algorithmic}[1]
\Function{Move}{}
\State Choose random index $i$ in $P$
\State Choose random index $j$ from the same partition to which $i$ belongs
\State Swap rows $i$ and $j$ in $P$
\EndFunction
\end{algorithmic}
\end{algorithm}
A partition is either the set of instance nodes at a single level in the STG (e.g. the set of functional harmonic key nodes), or the set of prototype nodes for a given feature (e.g. chord quality). Permuting only within valid partitions leverages the STG's inherent structure to avoid invalid moves globally detrimental to Equation \ref{eq:alignment_annealer_energy}. 

We set the alignment annealer's initial state to $P=I$, the identity matrix. By running the annealer for sufficiently many steps, we obtain optimal $P$.

\section{Centroid Derivation}\label{sec:centroid}
Centroid STG derivation is a constrained version of the Generalized Median Graph (GMG) problem, which, given a set of graphs, seeks to construct a prototype graph minimizing the distances over the input set graphs \cite{Blumenthal2021}. Formally, given a corpus of graphs $C$, the GMG $g$ is:
\begin{equation}
\label{eq:gmg}
g = \arg\min_{\hat{g}} \sum_{G \in C} d(\hat{g}, G)
\end{equation}
where $d$ is the distance measure. GMGs have wide applications in representation-based learning, particularly in biological settings \cite{Mukherjee2009}. Prior attempts at the GMG use genetic search \cite{Jiang2001}, linear programming\cite{Mukherjee2009}, block coordinate updates \cite{Blumenthal2021}, and median graph shift clustering \cite{Jouili2010}. To our knowledge, GMGs have never been applied to the music domain. Centroid STG derivation also operates in a significantly more constrained search space than the canonical GMG, since the centroid must be a well-formed STG. We thus propose a new approach combining nested simulated annealing with SMT solving.

\subsection{Bi-Level Simulated Annealing}\label{subsec:simanneal2}
Given padded adjacency matrix $A_g$ for an augmented centroid STG and its associated corpus of matrices $C = \{A_G\}$ that are optimally aligned to $A_g$, we seek to minimize $\textsc{Loss}$ in Equation \ref{eq:loss}, where $\textsc{Dist}$ is as in Equation \ref{eq:dist}. 
\begin{equation}
\textsc{Loss}(A_g, C) = \frac{1}{|C|} \sum_{A_G \in C} \textsc{Dist}(A_g, A_G)
\label{eq:loss}
\end{equation}

\noindent The centroid annealer's energy function is given in Equation \ref{eq:centroid_annealer_energy}, where $C_{\text{aligned}}$ is the corpus aligned to the current centroid $A_g$, a process itself requiring SA as in Section \ref{subsec:SA_align} to obtain the optimal alignments. Finding the centroid $A_g$ minimizing Equation \ref{eq:loss} is thus a nested NP-hard problem (GED and minimizing over these distances) requiring nested SA. 
\begin{equation}
\label{eq:centroid_annealer_energy}
\textsc{energy}(A_{G_1}, A_{G_2}, P) = \textsc{loss}(A_g, C_{\text{aligned}})
\end{equation}
As the centroid annealer's temperature cools, the loss converges as the centroid is increasingly closely aligned to its corpus. Thus, as the \textit{centroid} annealer's temperature cools, we scale down the number of steps and max temperature of the nested \textit{alignment} annealer.\footnote{See Appendix \ref{appendix_sec:sa_cooling} for precise cooling schedule}

The centroid annealer's move function for modifying the centroid $A_g$ at each step of SA is given in Algorithm \ref{alg:centroid_annealer_move}.
\begin{algorithm}[t]
\caption{Centroid Annealer Move Function}
\label{alg:centroid_annealer_move}
\begin{algorithmic}[1]
\Function{Move}{}
    \State Calculate the list of absolute difference matrices $D_L = [|A_g - A_{G_i}| \text{ for } A_{G_i} \in C_{aligned}]$
    \State Calculate the element-wise sum-of-distances score matrix. Higher score at coord $(i,j)$ means that coord has a higher impact on the loss
    \State Flatten $X$ and sort in descending score order 
    \State Partition $X_{\text{flat}}$ by unique score, and shuffle each partition randomly (increases variability of moves)
    \State Iterate through the indices $(i,j)$ of the sorted, partition-shuffled $X_{\text{flat}}$. Stop at the first (highest score) $(i,j)$ such that flipping the $(i,j)$ edge in $A_g$ is not:
        \begin{itemize}
        \item a globally structurally invalid move
        \item a move undoing the most recently accepted move (avoid oscillating states)
        \item a move the annealer has already locally rejected since the last accept (avoid getting stuck)
        \end{itemize}
    \State Execute move: $A_g[i,j] = 1 - A_g[i,j]$
\EndFunction
\end{algorithmic}
\end{algorithm}
To move strategically, we build the score matrix $S$ revealing which edge(s) in $A_g$ contribute most to the loss. We add or remove the edge at a highest score coordinate meeting the criterion in Algorithm \ref{alg:centroid_annealer_move}. In particular, a globally structurally invalid move induces a terminally invalid structure in the centroid by violating one of the rules in Table \ref{tab:global_constraints}.
\begin{table}[t]
\newcounter{globalitemnum}
\begin{tabular}{|p{0.01\linewidth} p{0.9\linewidth}|}
\hline
    \stepcounter{globalitemnum} \theglobalitemnum. & No self-loops \\
    \stepcounter{globalitemnum} \theglobalitemnum. & No instance-prototype or prototype-prototype edges\\
    \stepcounter{globalitemnum} \theglobalitemnum. & No edges from a prototype to an instance whose feature set does not include the proto feature (e.g. melody interval sign proto-segmentation instance) \\
    \stepcounter{globalitemnum} \theglobalitemnum. & No edges from lower to higher level instance levels (must respect the hierarchy) \\
    \stepcounter{globalitemnum} \theglobalitemnum. & No edges between non-adjacent instance levels (must respect k-partite structure)\\
    \hline
\end{tabular}
\caption{Global Constraints} \label{tab:global_constraints}
\end{table}
Some locally invalid moves, however, such as removing an edge in an intra-level linear chain, must be allowed as intermediate steps to a more optimal structurally valid state. Importantly, the STGs being compared must have the same number of levels; otherwise, edges spanning multiple levels must be allowed as they can be intermediate states towards the deletion of an entire level. Based on our experiments, this would be unacceptably detrimental to the performance of the annealer.

We set the centroid annealer's initial state $A_g$ to the corpus STG in the corpus with the min loss over the rest of the corpus. Running the annealer for sufficient steps gives an approximate centroid that may contain locally invalid states. 

\subsection{Graph Repair with SMT Solving}\label{subsec:z3}
In order to obtain a structurally sound centroid, we must ``repair" the approximate centroid $A_g$ by projecting it to the nearest valid STG. We achieve this by encoding the STG's structure as constraints in quantifier-free first-order logic formulae in the SMT (satisfiable modulo theory) solver Z3~\cite{z3}, which gives us formal guarantees on the soundness of the centroid. We use Z3's optimizer to minimize an objective over the constraints. Given approximate centroid $A_g$ and valid centroid $A_g'$, our objective is Equation \ref{eq:z3_obj}.
\begin{equation}
\label{eq:z3_obj}
\textsc{obj}(A_g, A_g') = \sum_i \sum_j |A_{g_{ij}} - A'_{g_{ij}}|
\end{equation}
Our constraints include the global rules in Table \ref{tab:global_constraints}, as well as additional constraints for instance nodes in Table \ref{tab:instance_constraints}, and finally prototype nodes in Table \ref{tab:proto_constraints}. We model relationships between nodes with uninterpreted functions.
\begin{table}[t]
\newcounter{institemnum}
\begin{tabular}{|p{0.01\linewidth} p{0.9\linewidth}|}
\hline
    \stepcounter{institemnum} \theinstitemnum. & Every instance node must have 1 or 2 instance parents in the level above \\
    \stepcounter{institemnum} \theinstitemnum. & The instances nodes at level $l$ must form a linear chain/total ordering via intra-level edges \\
    \stepcounter{institemnum} \theinstitemnum. & The start and end nodes of the linear chain must have the previous level linear chain's start and end nodes, respectively, as parents. \\
    \stepcounter{institemnum} \theinstitemnum. & In instance levels with non-overlapping nodes,\tablefootnote{Segmentation, keys, chords, and melody, but not motifs} the \emph{first} parent of a node at linear chain index $i > 0$, must not come before node $i-1$'s \emph{last} parent in the previous instance level's linear chain \\
    \stepcounter{institemnum} \theinstitemnum. & The \emph{first} parent of an instance node at linear chain index $i > 0$, must not come before node $i-1$'s \emph{first} parent in the previous level's linear chain \\
    \hline
\end{tabular}
\caption{Instance Constraints} \label{tab:instance_constraints}
\end{table}

\begin{table}[t]
\newcounter{protoitemnum}
\begin{tabular}{|p{0.01\linewidth} p{0.9\linewidth}|}
\hline
    \stepcounter{protoitemnum} \theprotoitemnum. & Every instance node must have exactly one prototype parent per feature \\
    \stepcounter{protoitemnum} \theprotoitemnum. & For levels that require it,\tablefootnote{Segmentation, keys, and chords only} no two linearly adjacent instance nodes can have identical prototype parent sets\\
    \hline
\end{tabular}
\caption{Prototype Constraints} \label{tab:proto_constraints}
\end{table}

Z3's optimizer supports integration with large neighborhood search (LNS) and can return intermediate semi-optimized solutions after a timeout. We run the optimizer with LNS, with initial soft constraints set to the approximate centroid $A_g$ to guide the optimizer. Even so, naively running the optimizer on a full STG is generally intractable due to combinatorial explosion, so we partition $A_g$ into subsets we can apply the constraints to incrementally. 

We first partition $A_g$ into pairs of consecutive instance levels without their prototypes (e.g. a segmentation/motif pair of instance levels), and optimize the instance constraints in Table \ref{tab:instance_constraints} and relevant global constraints in Table \ref{tab:global_constraints} over each partition incrementally. We combine the results of each partition until we build a valid centroid subgraph of instance nodes. Then, we partition $A_g$ into single levels, each containing the instance nodes of that level and all prototype nodes for each instance feature at that level (e.g. segmentation instance nodes + section number prototype nodes). The instance constraints are already optimized; we need only optimize the prototype constraints in Table \ref{tab:instance_constraints} and relevant global constraints in Table \ref{tab:global_constraints} over the possible prototypes. This gives us the complete, structurally sound centroid $A'_g$ we seek.\footnote{See Appendix \ref{appendix_sec:centroid_derivation} for centroid derivation visualization}

\section{Experiments}\label{sec:eval}
We conduct experiments to verify that: (1) structural distance accurately differentiates individual pieces, with its performance reliant on the complete hierarchy, and (2) the centroid encapsulates the overarching structure of its corpus.

To evaluate our approach, we create a dataset of polyphonic, symbolic MIDI piano music from the 
\href{https://www.kunstderfuge.com/}{Kunstderfuge} and \href{http://www.piano-midi.de/}{Classical Piano MIDI} datasets.
Since some single-level analyses we use to generate STGs require the data to be in audio and CSV format, we convert MIDI to CSV (with a manual script) and to MP3 (with Fluidsynth). 

We then generate an STG for each piece. For segmentation, we use the flat Structural Features algorithm~\cite{sf} for segment boundaries and 2D-Fourier Magnitude Coefficients~\cite{2dfmc} for segment labels, both of which are provided by the Music Structure Analysis Framework~\cite{msaf}. For motifs, we use the BPS-motif discovery algorithm~\cite{Hsiao_2023_motifs}, and for relative keys and functional harmonic chords we use the pretrained Harmony Transformer V2~\cite{harmonytransformerV2}. Finally, for melodic contour, we use the Melodia algorithm~\cite{salamon_2013_melody}.

\subsection{Structural Distance Evaluation} \label{subsec:eval_struct_dist}
\begin{figure}[t]
  \centering
  \includegraphics[width=\linewidth]{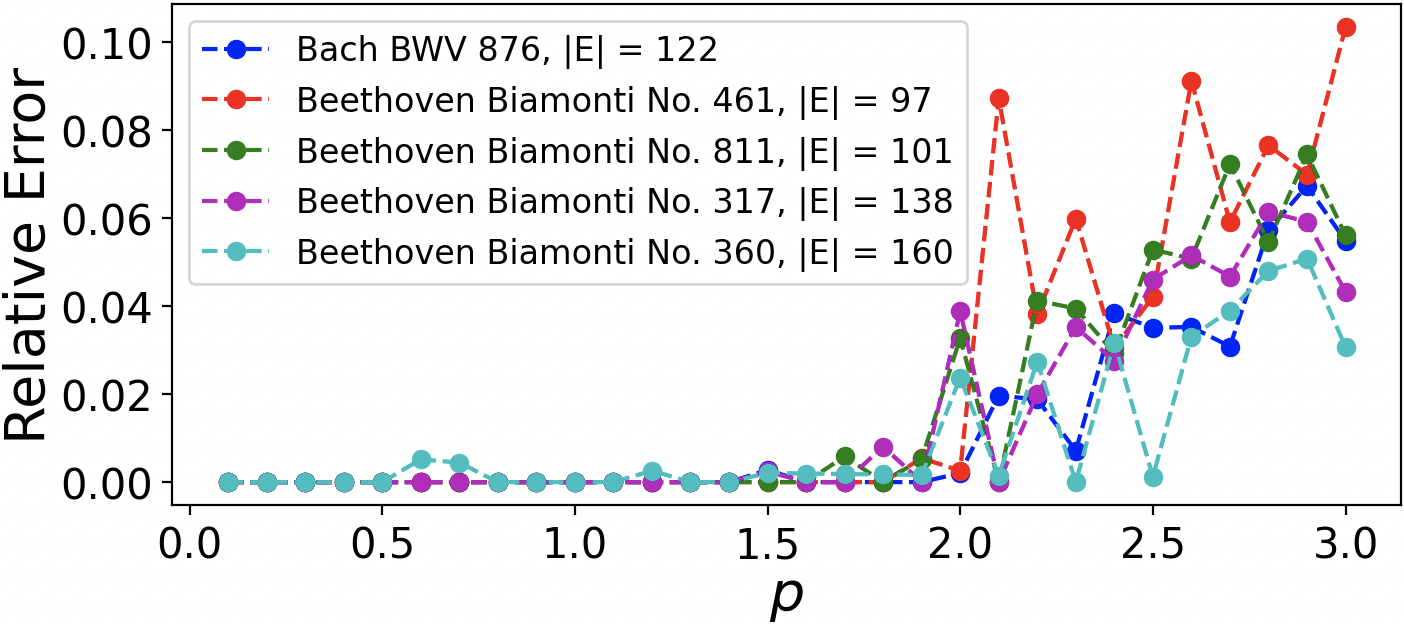}
  \caption{Relative Error: Computed vs Ground-Truth Structural Dist}
  \label{fig:synthetic_struct_dist}
\end{figure}

\noindent\textbf{Mathematical Verification.} 
Given two STGs $G_1, G_2$ recall that under optimal alignment, $\textsc{dist}(A_{G_1}, A_{G_2})$ is $\sqrt{\text{GED}}$. To evaluate our alignment annealer, we set $G_1$ with $|E_1|$ edges as the ``base graph." From $G_1$, we generate a series of STGs $G_2$ by randomly adding $\ceil{|E_1|\cdot p}$ valid edits to $G_1$ (add/remove edge, verified with Section \ref{subsec:z3} Z3 solver), where $p \in \{0.1, 0.2, \ldots, 3\}$ (i.e. $p$ ranges from 10-300\% edits in the size of $|E_1|$). We evaluate structural distance as a function of $p$ for five base STGs $G_1$ by computing the relative error from experimental $\textsc{dist}(A_{G_1}, A_{G_2})$ to ground truth structural distance $\sqrt{\ceil{|E_1|\cdot p}}$ (Figure \ref{fig:synthetic_struct_dist}). Relative error is close to 0 (perfect alignment) for $p<1.8$, and only deteriorates, at worst, to 10.33\% at $p=3$ for the Beethoven 461 base graph.

\smallskip\noindent\textbf{Musical Evaluation.} To verify structural distance accurately differentiates pieces, we construct 210 STG sets from 32 pieces by J.S. Bach (21), Mozart (2), Beethoven (3), Schubert (2), and Chopin (4). Each set is a unique combination of 5 pieces, one from each composer, such that a piece's duration is within 7 seconds of any other piece in the set, since structural distance between disparate length pieces could be due to STG size differences rather than the local structural variations we aim to distinguish.\footnote{See Appendix \ref{appendix_sec:input_pieces} for input pieces/sets details} We compute pairwise structural distances between the STGs in each set with our Section \ref{sec:struct_dist} graph alignment annealer (2000 iterations, max and min temperature of 2 and 0.01) on 8 Nvidia RTX 2080 GPUs with 11GB RAM. This results in 210 structural distance matrices, which we average to a single mean distance matrix.

We evaluate structural distance against three baselines over the same 210 piece combinations. Baseline 1 is the mean distance matrix obtained by taking the cosine similarity between feature vectors extracted from each MIDI file using Music21. Baseline 2 is the mean distance matrix over pairwise Stent weighted audio similarities (SWAS) for each piece's paired MP3 file. SWAS is a composite audio similarity metric comprising zero-crossing rate, rhythm, chroma, spectral contrast, and perceptual similarity metrics, which we weigh equally. To demonstrate existing graph comparison metrics are insufficient, Baseline 3 is the Weisfeiler-Lehman (WL) Kernel applied pairwise to the STGs in each set, with five iterations and Neighborhood Hash (NH) as the base kernel. The WL Kernel iteratively refines node labels based on their neighbors, and uses a base kernel (in this case, NH, which efficiently captures local graph structure by distinguishing neighborhood configurations) to compare them \cite{wl_kernel}. Our ground-truth is the stylistic similarity indices between composers using human annotations and metadata from ``The Classical Music Navigator"~\cite{classicalmusicnavigator}.

We normalize all matrices to [0, 1] range and apply the Mantel test with Spearman's rank correlation coefficient $\rho_s$ to evaluate our results against the ground-truth similarity from the Classical Music Navigator (Table \ref{tab:struct_dist_results}). $\rho_s$ is highest (with $p < 0.05$) for structural distance (SD in Table \ref{tab:struct_dist_results}), verifying that structural distance accurately differentiates between pieces and captures human conception of musical similarity.
\begin{table}[t]
    \centering
    \begin{tabular}{l|cc}
    \toprule
        Metric & $\rho_s$ & $p$-value \\\midrule 
        \textbf{Ours: SD} & \textbf{0.8207} & \textbf{0.0130} \\
        Baseline 1: MIDI Features & 0.4681 & 0.3150 \\
        Baseline 2: SWAS & 0.5775 & 0.1690 \\
        Baseline 3: WL Kernel (+NH base) & -0.8389 & 0.0110\\
        \bottomrule
    \end{tabular}
    \caption{Mantel Test with Spearman's rank correlation coefficient for normalized mean distance matrices}
\label{tab:struct_dist_results}
\end{table}

Finally, to verify the importance of the full hierarchy, we repeat this experiment with bottom-up level ablations of the STG (Table \ref{tab:struct_dist_ablation}). The structural distance algorithm's performance incrementally declines each time a layer of the STG is removed, confirming the necessity of the complete hierarchy.

\begin{table}[t]
    \centering
    \begin{tabular}{l|cc}
    \toprule
        Metric & $\rho_s$ & $p$-value \\\midrule 
        \textbf{SD - complete STG (5 levels)} & \textbf{0.8207} & \textbf{0.0130} \\
        SD - 4 levels & 0.7842 & 0.0390 \\
        SD - 3 levels & 0.7173 & 0.0680 \\
        SD - 2 levels & 0.6930 & 0.1150 \\
        SD - 1 level & -0.4377 & 0.2810 \\
        \bottomrule
    \end{tabular}
    \caption{Mantel Test with Spearman's rank correlation coefficient for normalized mean distance matrices with STG level ablations (first row is same as Table \ref{tab:struct_dist_results})}
\label{tab:struct_dist_ablation}
\end{table}

\subsection{Centroid Evaluation} \label{subsec:eval_centroid}
\noindent\textbf{Mathematical Verification.} As a correctness check, we evaluate derived centroids against constructed ground truth centroids. Given base STG $g$ with $|E|$ edges, we create a synthetic corpus $C_k$ of $k$ STGs by randomly adding $\ceil{\frac{|E|}{2}}$ valid edits to $g$, $k$ times, as in Section \ref{subsec:eval_struct_dist}. By construction, $g$ must be a true centroid with min possible loss over $C_k$ (Equation \ref{eq:loss}).\footnote{Proof in Appendix \ref{appendix_sec:proof_centroid}} We derive candidate centroid $g_d$ from $C_k$, and compare to $g$. We evaluate $g_d$ against the naive centroid $g_n$, \emph{i.e.,} the STG already in $C_k$ with min loss over the rest of $C_k$.

We choose our base STG $g$ to be Beethoven's Biamonti Sketch No. 461, which, when augmented, has $|E|=97$ edges. Thus, we add $\ceil{\frac{97}{2}}=49$ edits to $g$, $k$ times, to generate a synthetic corpus $C_k$ of size $k$ with $g$ as its true centroid. 49 edits is structural distance of 7.0, since structural distance between optimally aligned graphs is $\sqrt{\text{GED}}$ (Equation \ref{eq:dist}). 

We repeat this process for $k\in[3,14]$. For each corpus $C_k$, we use the same GPU infrastructure as Section \ref{subsec:eval_struct_dist} to generate approximate centroids with the centroid annealer (1000 iterations, max and min temperature of 2.5 and 0.05). At each iteration, we run the nested graph alignment annealer, starting at 500 steps, max and min temperature of 1 and 0.01, and ending at 5 steps, max and min temperature of 0.05 and 0.01, as the outer centroid annealer's loss converges (see Section \ref{subsec:simanneal2}). We run our Z3 optimizer to generate each final, repaired centroid $g_d$, using 24 Intel Xeon cores and 32GB RAM. 

\begin{figure}[t]
  \centering
  \includegraphics[width=\linewidth]{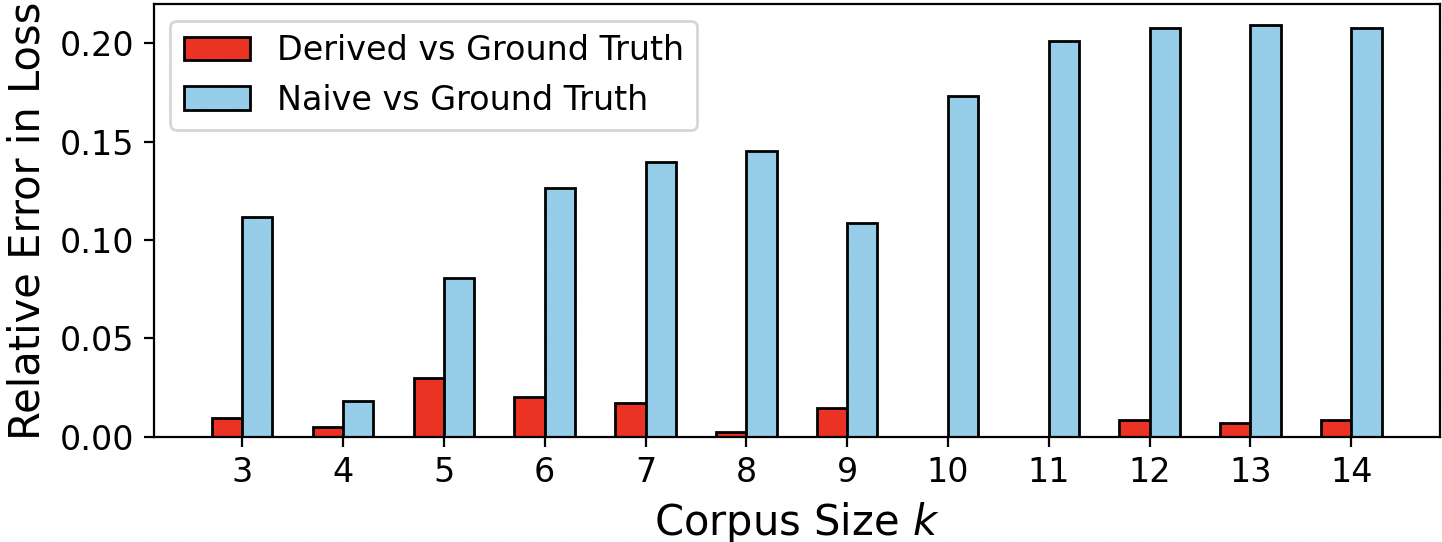}
  \caption{Relative Error in Loss ([0,1] range): Derived and Naive Centroids vs Ground-Truth Centroid}
  \label{fig:synthetic_centroid_loss_err}
\end{figure}

There may be multiple non-isomorphic true centroids with the same minimal loss over $C_k$. Evaluating the relative error between optimal loss from $g$ and experimental loss from $g_d$ thus gives a more rigorous assessment of $g_d$ than directly comparing $g_d$ to $g$. Figure \ref{fig:synthetic_centroid_loss_err} shows the relative error in loss $E^{g}_{g_d}$ between $g$ and $g_d$ for $k \in [3,14]$ in red, compared to the relative error in loss $E^{g}_{g_n}$ between $g$ and $g_n$ in blue. We observe $E^{g}_{g_d}$ is consistently small, with max $E^{g}_{g_d}= 2.99\%$ at $k=5$, and min $E^{g}_{g_d}=0$ (i.e. $g_d$ is a true centroid) at $k=10,11$. $g_d$ also greatly outperforms $g_n$: for nonzero $E^{g}_{g_d}$, $E^{g}_{g_n}$ is on average 17.23 times worse than $E^{g}_{g_d}$.

\smallskip \noindent\textbf{Musical Evaluation.}  
This experiment requires pieces with more length (and STG size) variation than Section \ref{subsec:eval_struct_dist}, as centroids derived from very similar input graphs may be trivial by construction and fail to generalize to more diverse corpora. We thus relax the Section \ref{subsec:eval_struct_dist} relative duration restriction to 80 seconds, and generate centroid STGs for four corpora with pieces by Alkan (11), Chopin (8), Haydn (12), and Mozart (14) with the same architecture as before.\footnote{See Appendix \ref{appendix_sec:input_pieces} for input pieces details}

\begin{figure}[t]
  \centering
  \includegraphics[width=0.35\linewidth]{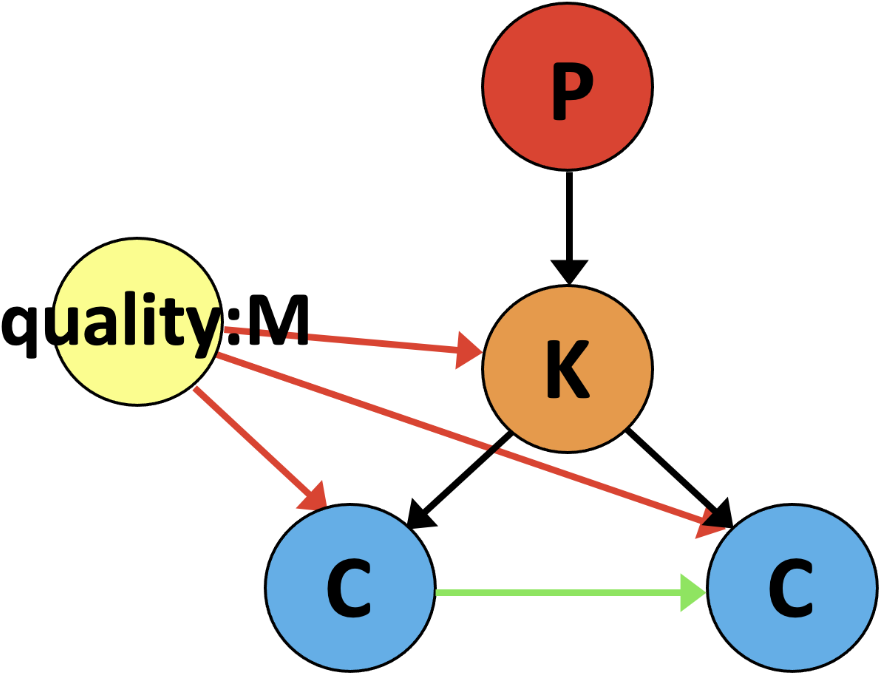}
  \caption{Example Common 5-Node Subgraph of Mozart Corpus}
  \label{fig:frequent_subgraph}
\end{figure}

To verify each derived centroid $g$ musically characterizes its corpus $C$, we use rustworkx \cite{rustworkx} to enumerate the set of all 5-node subgraphs ($\{S_5\}$) common to every STG in $C$, thus extracting the most structurally salient musical relationships in $C$.\footnote{Larger subgraphs encode more robust relationships, but mining $k$-node subgraphs for $k>5$ was intractable, and approximate subgraph mining tools were too imprecise for meaningful conclusions} For each $C$, we evaluate the percentage of the graphs in $\{S_5\}$ that are also subgraphs of $g$; i.e. we evaluate how well $g$ captures the most structurally salient musical relationships in $C$ (Table \ref{tab:substructure_freq}). On average, $70.33\%$ of each $\{S_5\}$ aligns \emph{perfectly} with $g$, confirming each centroid captures the musically essential information of its corpus. To visualize these essential musical substructures, consider a 5-node subgraph common to the Mozart corpus and captured in its centroid (Figure \ref{fig:frequent_subgraph}). This reveals that a characterizing feature of the Mozart corpus is two consecutive major chords in the same major key, which in turn falls in a motif/pattern.

\begin{table}[t]
    \centering
    \begin{tabular}{l|c|c}
    \toprule
        Composer & Size of $\{S_5\}$ & \% in Derived Centroid \\\midrule 
        Alkan & 1804 & 76.67\% \\
        Chopin & 2111 & 70.06\% \\
        Haydn & 2504 & 63.90\% \\
        Mozart & 1900 & 70.68\% \\
        \bottomrule
    \end{tabular}
    \caption{Analysis of Common 5-Node Subgraphs $\{S_5\}$}
\label{tab:substructure_freq}
\end{table}

\section{Conclusion and Future Work}
We presented the \emph{structural temporal graph} (STG) to encapsulate complete, hierarchical musical structure; a measure of \emph{structural distance} between STGs; and an algorithm to derive a \emph{centroid STG} structurally representing a music corpus. We showed structural distance and derived centroids both mathematically approximate ground truth; structural distance accurately differentiates music pieces; and derived centroids capture the essential structural relationships of their corpora.

The STG and derived centroids lay the groundwork for structured, controllable sequence data generation. For example, users could modify the STG of a generated music piece to update constraints on a generative model.  Beyond music, the STG can model structural hierarchies for \emph{any} sequence data given algorithms for analyses at each level. For instance, an STG could encode a poetry hierarchy---verses, stanzas, lines---with its centroid structurally summarizing the poetry corpus. Such applications enable human refinement of machine-generated data to meet desired structural specifications across fields.

\appendix
\section{Nested Alignment Annealer Cooling Schedule in Centroid Annealing}
\label{appendix_sec:sa_cooling}
Recall that as the centroid annealer's temperature cools, the loss converges as the centroid is increasingly closely aligned to its corpus. Thus, as the \textit{centroid} annealer's temperature cools, we scale down the number of steps and max temperature of the nested \textit{alignment} annealer. Specifically, given the hyperparameters $T_{\text{initial\_max}}$ (initial max temperature), $T_{\text{final\_max}}$ (final max temperature), $S_{\text{initial}}$ (initial number of steps), and $S_{\text{final}}$ (final number of steps) of the nested alignment annealer, its current temperature $T_{\text{curr}}$ and current number of steps $S_{\text{curr}}$ are as follows:

\begin{itemize}
\item [(1)] \begin{align*}
R = \frac{T_{\text{curr}}^c - T_{\text{min}}^c}{T_{\text{max}}^c - T_{\text{min}}^c}
\end{align*}
where $T_{\text{curr}}^c,T_{\text{min}}^c$, and $T_{\text{max}}^c$ are the current, min, and max temperatures of the outer centroid annealer. $R$ measures the normalized cooling ratio of the centroid annealer, i.e. the proportion of cooling that is complete.

\item [(2)]
\begin{align*}
T_{\text{curr}}&=T_{\text{initial\_max}} \cdot R + T_{\text{final\_max}}\cdot (1-R)
\end{align*}

\item [(3)]
\begin{align*}
S_{\text{curr}}&=\lfloor(S_{\text{initial}}\cdot R + S_{\text{final}}\cdot (1-R))\rfloor
\end{align*}
\end{itemize}

\section{Guided Example of Centroid Generation}
\label{appendix_sec:centroid_derivation}
\begin{figure*}[p]
  \centering
  \includegraphics[width=\linewidth, height=0.92\textheight, keepaspectratio]{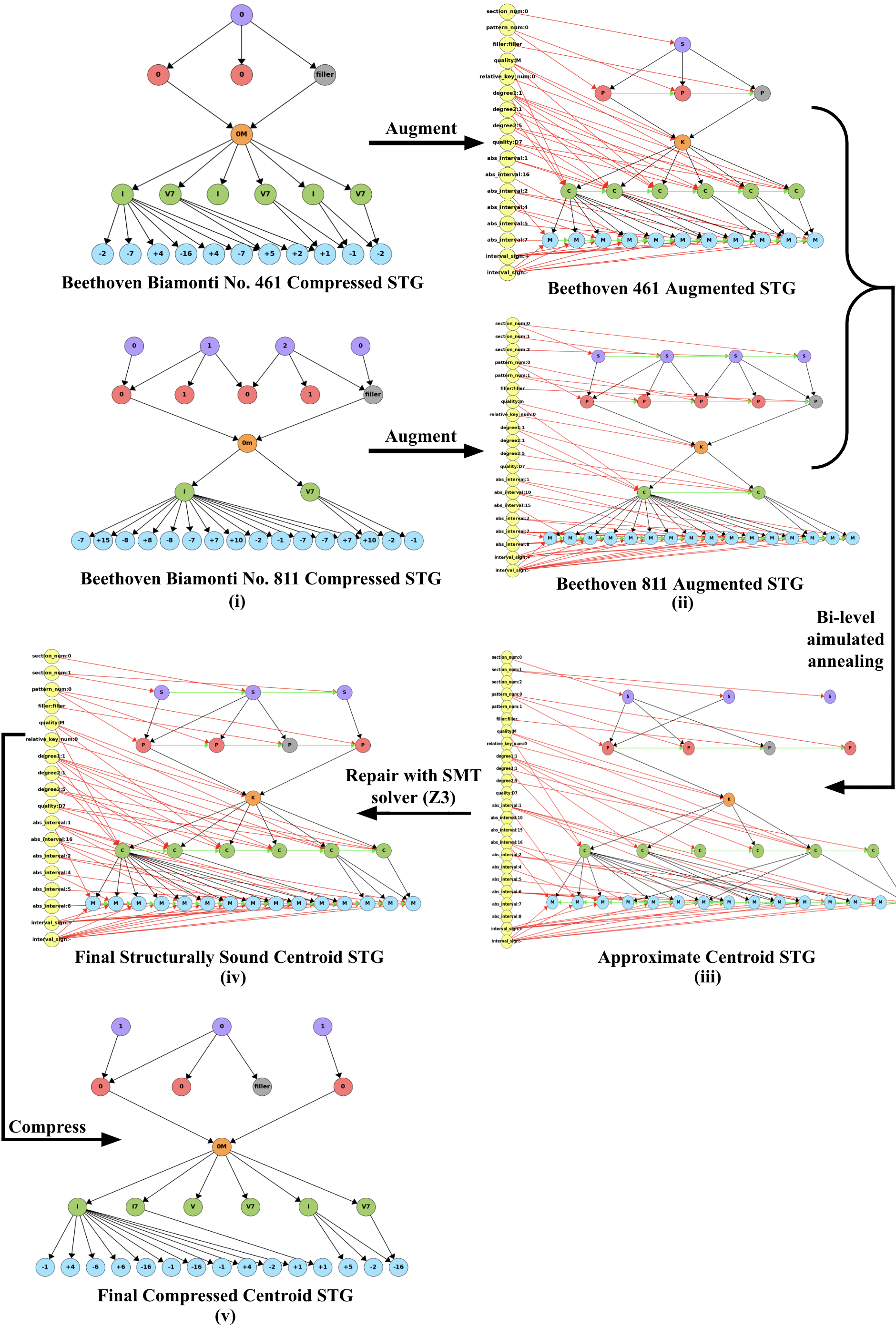}
  \caption{Example centroid derivation. In each STG, the segmentation nodes are purple, the motif nodes are red, the filler nodes are gray, the key nodes are orange, the chord nodes are green, and the melody nodes are blue. On the augmented STGs in (ii)-(iv), prototype nodes are to the left in yellow, prototype-instance edges are red, and intra-level linear chain edges are in green. Table \ref{tab:proto_legend} explicates the prototype node labels in more detail. Refer to Figure \ref{fig:beethoven_stg_aug_zoom} in the main text for a zoomed-in view of the prototype nodes.}
  \label{fig:centroid_derivation}
\end{figure*}

To better comprehend the centroid derivation process in Sections \ref{sec:struct_dist} and \ref{sec:centroid}, we walk through the five steps of an example centroid derivation from a corpus of two STGs in Figure \ref{fig:centroid_derivation}, found at the bottom of this appendix. Each of the following steps corresponds to the analogously labeled step in Figure \ref{fig:centroid_derivation}.

\begin{enumerate}[label=(\roman*), left=0em]  % Lowercase Roman numerals
  \item We start with the corpus we will derive the centroid from. Our corpus contains two STGs. The first STG is the same as in Figure \ref{fig:beethoven_stg} in the main text: Beethoven's Biamonti Sketch No. 461. The second STG is from Beethoven's Biamonti Sketch No. 811. Both STGs were built using the process in Section \ref{sec:stg}.
  
  \item We augment both STGs from (i) as in Section \ref{sec:struct_dist}. The implicit structure encoded in the node labels in (i) is now explicitly encoded in the graph's topology in (ii) in Figure \ref{fig:centroid_derivation} via the yellow prototype nodes, red prototype-instance edges, and green intra-level linear chains. The instance node labels thus are updated to only show the layer kind now (S for segmentation, P for patterns/motifs, K for keys, C for chords, or M for melody) Recall that each yellow prototype parent node encodes a \textit{feature} of its instance child, and each prototype label is of the form \verb!feature_name:feature_value!. The names and possible values of the features for each level of the STGs in Figure \ref{fig:centroid_derivation} are explicated in Table \ref{tab:proto_legend}.
  
  \item Since all structure is now explicit and we can reason about the graphs isomorphically, we can now apply the nested simulated annealing procedure from Sections \ref{subsec:SA_align} and \ref{subsec:simanneal2} to obtain the approximate centroid STG for the corpus. Some locally invalid states remain, such as the lack of the intra-level linear chain in the purple segmentation level.

  \item We arrive at the final, well-formed centroid STG by projecting the approximate centroid STG from (iii) onto the nearest valid STG as in Section \ref{subsec:z3}, using the SMT solver Z3 to impose guarantees of structural soundness on the result. Notice, for instance, how the requisite linear chain has been added to the purple segmentation level. 
  
  \item Finally, we convert the repaired centroid from (iv) back to its compressed form (i.e. we move the information encoded in the prototype nodes and edges and the intra-level linear chains, back into the nodes labels), which allows us to visually compare it to the original STGs in the corpus from (i).
\end{enumerate}

\begin{table}[t]
\centering
\resizebox{\columnwidth}{!}{
    \begin{tabular}{|l|l|l|}
    \hline
    \textbf{Level}        & \textbf{Feature Name}       & \textbf{Feature Values}                                         \\ \hline
    Segmentation & \texttt{section\_num}       & $\mathbb{Z}_{\geq 0}$ \\ \hline
    Motifs       & \texttt{pattern\_num}       & $\mathbb{Z}_{\geq 0}$ \\ \hline
    Motifs       & \texttt{filler}           & filler \\ \hline
    Keys         & \texttt{relative\_key\_num} & $\mathbb{Z}_{\geq 0}$ \\ \hline
    Keys         & \texttt{quality}            & \{M, m\}                                               \\ \hline
    Chords       & \texttt{quality}            & \{M, m, d, d7, h7, \\&&\;D7, a, a6, a7\}                                               \\ \hline
    Chords       & \texttt{degree1}            & $\{1, 2,\ldots, 12\}$                                      \\ \hline
    Chords       & \texttt{degree2}            & $\{1, 2, \ldots, 12\}$                                      \\ \hline
    Melody       & \texttt{abs\_interval}      & $\mathbb{Z}$                            \\ \hline
    Melody       & \texttt{interval\_sign}     & $\{+, -\}$                                               \\ \hline
    \end{tabular}
}
\caption{Prototype feature names (middle) and possible values (right) for each instance level (left). The chord quality values are shorthand for major, minor, augmented, diminished, half-diminished, dominant, etc. degree1 and degree2 refer to the primary and secondary degrees, respectively, in the diatonic scale that the chord is built on. Putting it all together, an example prototype label could be degree1:1, quality:M, filler:filler, or abs\_interval:5.}
\label{tab:proto_legend}
\end{table}

\section{Input Pieces for Section \ref{sec:eval} Experiments}
\label{appendix_sec:input_pieces}
For details about the pieces used in the experiments, please consult our \href{https://github.com/ilanashapiro/stg_optimization}{GitHub repository}.\footnote{\url{https://github.com/ilanashapiro/stg\_optimization}} Here, the 32 pieces used for the Structural Experiment (Section \ref{subsec:eval_struct_dist}) can be found in the file:
\renewcommand{\UrlFont}{\ttfamily}
\url{project/experiments/structural\_distance/structural\_distance\_experiment/input\_pieces.txt}.
The 210 set combinations of these pieces used in this experiment can be found in:
\url{project/experiments/structural\_distance/structural\_distance\_experiment/set\_combinations.txt}.
Finally, the pieces used for the Centroid Experiment (Section \ref{subsec:eval_centroid}) can be found in the file:
\url{project/experiments/centroid/substructure\_frequency\_experiment/corpora/info.txt}.

\section{Proof $g$ is True Centroid of Synthetic Corpus}
\label{appendix_sec:proof_centroid}

Recall that from STG $g$ with $|E|$ edges, we generate a synthetic corpus of size $k$ by randomly adding $n = \ceil{\frac{|E|}{2}}$ edits, or edit operations (add or remove edge), to $g$. We repeat the edit addition $k$ times to build a synthetic corpus $C$ of size $k$, where each noisy STG is equidistant from $g$, and verify $C$ consists of valid STGs with our Z3 solver from Section \ref{subsec:z3}. We now prove why $g$ must be the true centroid of $C$.

Each graph in $C$ is generated by applying exactly $n$ edit operations to $g$, so for all $G \in C$, the graph edit distance (GED) $d(g, G) = n$. The sum of distances (SOD) from $g$ to all graphs in $C$ is therefore 

\[
\sum_{G \in C} d(g, G) = k \cdot n.
\]

Suppose, for contradiction, that there exists another graph $g' \neq g$ with a lower SOD. This would imply that
\[
\sum_{G \in C} d(g', G) < k \cdot n.
\]

Since GED is known to be metric \cite{ged}, it satisfies the triangle inequality, so for any $G \in C$, we have
\[
d(g', G) \geq |d(g', g) - d(g, G)|.
\]

Because $d(g, G) = n \; \forall G \in C$, it follows that
\[
d(g', G) \geq |d(g', g) - n|.
\]

If $g' \neq g$, then $d(g', g) > 0$, so the expected distance from $g'$ to a randomly chosen $G \in C$ is at least $n$. Thus, for any alternative $g' \neq g$, the total SOD
\[
\sum_{G \in C} d(g', G)
\]
\noindent cannot be lower than $k \cdot n$, contradicting the assumption that $g'$ has a lower SOD. 

Notably, GED forms a 1-1 correspondence with structural distance, since structural distance between optimally aligned graphs is simply the square root of GED by definition (Equation \ref{eq:dist}). Therefore, $g$ has minimal SOD, and is thus the true centroid. Notably, $g$ is not unique; there may be multiple non-isomorphic graphs with the same minimal SOD, but since they equally minimize SOD, they are considered equivalent as centroids by definition.

\section*{Acknowledgments}
This work is supported by National Science Foundation grants \#1955457, \#2220892, \#2146151, and \#2200333, and by the European Research Council under Europe's Horizon 2020 program, grant \#883313 (ERC REACH).

\bibliographystyle{named}
\bibliography{main}

\end{document}